\documentclass{article}
\usepackage{ijcai25}
\usepackage{times}
\usepackage{subfigure}
\usepackage{tabularx}
\usepackage{pifont} 
\usepackage{soul}
\usepackage{url}
\usepackage[hidelinks]{hyperref}
\usepackage[utf8]{inputenc}
\usepackage[small]{caption}
\usepackage{graphicx}
\usepackage{amsmath}
\usepackage{amsthm}
\usepackage{booktabs}
\usepackage{algorithm}
\usepackage{algorithmic}
\usepackage{amssymb}
\usepackage{multirow}
\usepackage[switch]{lineno}
\title{Beyond Patterns: Harnessing Causal Logic for Autonomous Driving \\Trajectory Prediction}
\author{
Bonan Wang\textsuperscript{\rm 1}\thanks{Authors contributed equally; \dag Corresponding author.}\and
Haicheng Liao\textsuperscript{\rm 1}$^{*}$\and
Chengyue Wang\textsuperscript{\rm 1}\and
Bin Rao\textsuperscript{\rm 1}\and
Yanchen Guan\textsuperscript{\rm 1}\and\\
Guyang Yu\textsuperscript{\rm 1}\and
Jiaxun Zhang\textsuperscript{\rm 1}\and
Songning Lai\textsuperscript{\rm 2}\and
Chengzhong Xu\textsuperscript{\rm 1}\and
Zhenning Li\textsuperscript{\rm 1}$^{\dag}$
\\
\affiliations
$^1$University of Macau\\
$^2$The Hong Kong University of Science and Technology (Guangzhou)\\
\emails
\{mc35002, yc27979, yc47938, binrao, yc37976, yc47415, czxu, zhenningli\}@um.edu.mo, yuguyangsam@gmail.com,
songninglai@hkust-gz.edu.cn
}
\begin{document}
\maketitle
\begin{abstract}
    Accurate trajectory prediction has long been a major challenge for autonomous driving (AD). Traditional data-driven models predominantly rely on statistical correlations, often overlooking the causal relationships that govern traffic behavior. In this paper, we introduce a novel trajectory prediction framework that leverages causal inference to enhance predictive robustness,  generalization, and accuracy. By decomposing the environment into spatial and temporal components, our approach identifies and mitigates spurious correlations, uncovering genuine causal relationships. We also employ a progressive fusion strategy to integrate multimodal information, simulating human-like reasoning processes and enabling real-time inference. Evaluations on five real-world datasets—ApolloScape, nuScenes, NGSIM, HighD, and MoCAD—demonstrate our model's superiority over existing state-of-the-art (SOTA) methods, with improvements in key metrics such as RMSE and FDE. Our findings highlight the potential of causal reasoning to transform trajectory prediction, paving the way for robust AD systems.
\end{abstract}
\section{Introduction}
\begin{figure}[!ht]
\centering
\includegraphics[width=0.5\textwidth]{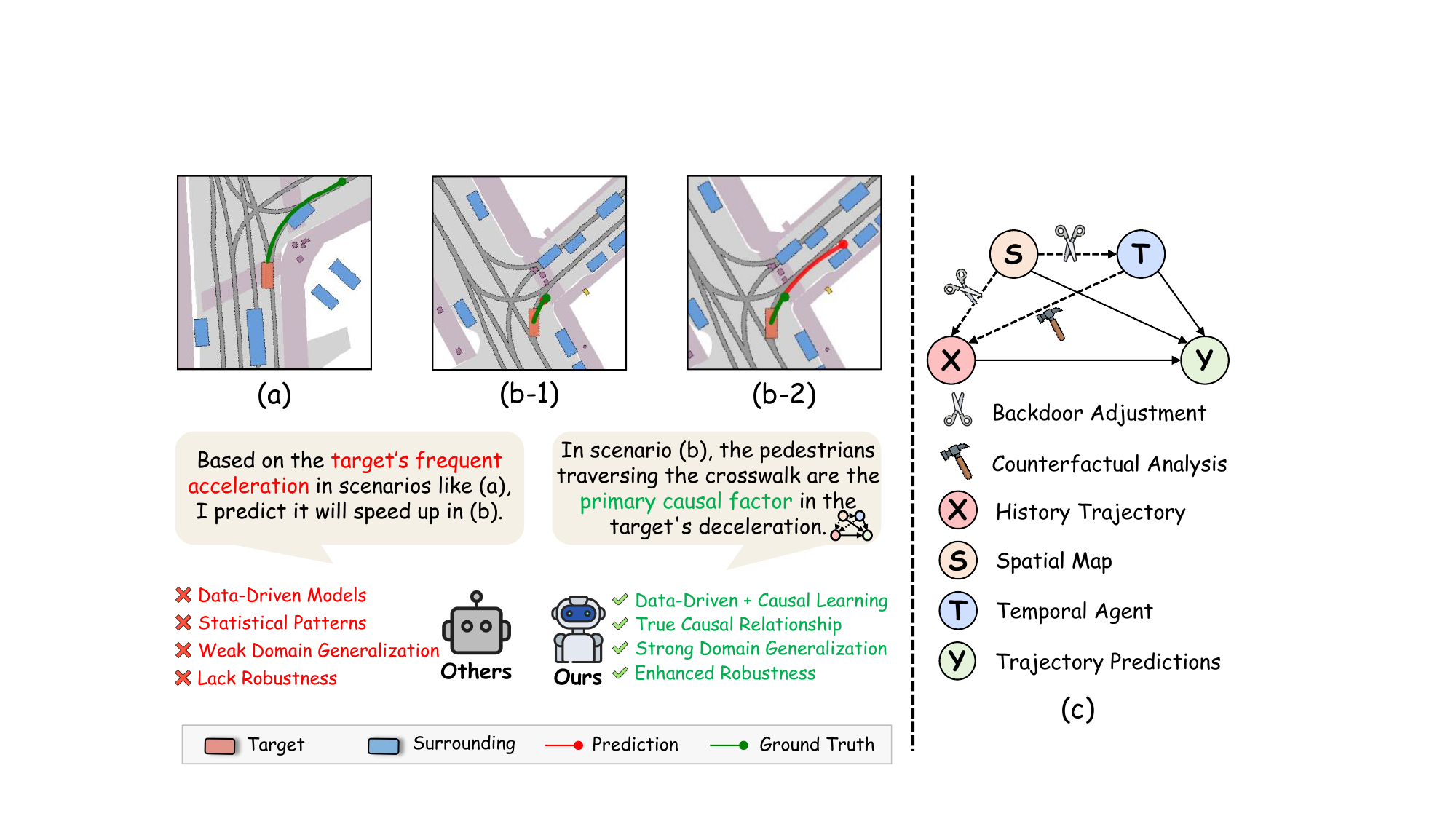}
\caption{
Illustration of causal relationships in traffic scenarios. Panel (a) represents the training stage, where the target agent is frequently observed accelerating through crosswalks. In the test stage, a traditional data-driven model predicts that the target agent will similarly accelerate, as shown in (b-1). In contrast, our model employs a causal inference to discern the true causal relationship, enabling the accurate prediction of the target agent's behavior of stopping at the crosswalk, as demonstrated in (b-2). Panel (c) shows the proposed new causal paradigm utilizes backdoor inference and counterfactual analysis to mitigate the confounding effects of spatial map $S$ and temporal agent $T$ data.
}
\label{fig:head}
\end{figure}
Autonomous driving (AD) technology holds promise for revolutionizing transportation by enhancing traffic efficiency and safety \cite{tsliao,guan2024world}. A critical component of AD systems is trajectory prediction, which involves forecasting the future positions of vehicles and other traffic participants. Accurate trajectory prediction enables these systems to anticipate and respond to dynamic changes in the environment, providing essential inputs for decision-making, route planning, and collision avoidance \cite{liao2024less}.
Despite significant advancements in trajectory prediction for autonomous vehicles (AVs), modern data-driven models primarily focus on identifying \textbf{statistical patterns} within large datasets \cite{chen2021human}. While this approach has been effective in capturing correlations, it often overlooks the deeper causal relationships that underpin driving behaviors. These models \cite{liao2024bat} tend to treat driving environments as mere data-generating processes without recognizing the causative factors that influence vehicle dynamics.

As illustrated in Figure \ref{fig:head}, the model may learn that vehicles tend to accelerate in particular scenarios, such as scenario (a). This insight allows the model to anticipate acceleration maneuvers at analogous points in the future, as demonstrated in scenario (b). However, such predictions are based solely on observed patterns and lack an understanding of the underlying causes, such as the presence of crosswalks or sharp curves in the road. This superficial reliance on data correlations results in models that are tethered to the training data, without the capability to infer causal links, such as the impact of pedestrian movement (temporal agent data) or changes in road layout (spatial map data). These limitations become particularly problematic when the model encounters scenarios that are not well-represented in the training data. Without an understanding of causal mechanisms, the model struggles to generalize to new situations, potentially leading to inaccurate predictions and safety risks \cite{chen2021human,liao2025cot,ge2023causal}. This highlights the need for trajectory prediction methods that incorporate causal reasoning to better interpret and predict complex traffic behaviors. 

This observation highlights the need for a paradigm shift toward \textbf{causal inference} in trajectory prediction, emphasizing the understanding of cause-and-effect relationships within data. Such an approach is crucial for AD systems for several reasons. First, causal inference enhances the robustness of the model \cite{bagi2023generative}. AVs operate in dynamic scenes and may encounter sensor noise, missing data, or abnormal driving behaviors. By identifying underlying causal mechanisms, causal inference can effectively filter out these noises or random perturbations, maintaining the stability and accuracy of predictions. Second, causal inference improves generalization to unseen scenarios, which is essential for reliable performance in diverse environments \cite{ge2023causal,chen2021human}. Finally, causal models offer increased interpretability by providing insights into why specific predictions are made, thereby enhancing transparency and trust.

To address these needs, we propose a novel causal inference framework for trajectory prediction in AD. Causal inference is a method that goes beyond merely identifying correlations in data to uncover the actual cause-and-effect relationships that drive the behaviors being observed \cite{pearl2009causality,kuang2020causal}. This is particularly important in dynamic environments like traffic scenarios, where understanding the underlying causes of actions, rather than just patterns, leads to more accurate trajectory predictions for AVs. Our approach constructs a causal graph to explicitly represent the relationships between key variables, such as spatial map data and temporal agent data. As illustrated in Figure \ref{fig:head} (c), we employ a causal inference paradigm, including backdoor adjustment and counterfactual analysis, to isolate the confounding variables in the traffic environment. These paradigms eliminate spurious correlations, uncovering the true causal relationships in the data. Furthermore, we introduce a cross-modal progressive fusion strategy that mimics the gradual reasoning process of human drivers. This strategy incorporates a multi-stage attention mechanism to generate progressive causal queries to improve prediction accuracy.

In summary, our contributions are as follows:
\begin{itemize}
\item We propose a novel \textbf{causal inference} paradigm for trajectory prediction in AD, utilizing a causal graph to model cause-and-effect relationships between spatial map data and temporal agent data. This approach enables the identification of genuine causal links, improving prediction accuracy and robustness.
\item We introduce a \textbf{cross-modal progressive fusion} strategy, using a multi-stage attention mechanism to integrate multimodal information. This enhances real-time inference, allowing the model to adapt to dynamic scenes.
\item Our extensive experiments across five real-world datasets consistently demonstrate that both our causal inference model and its plug-in module outperform existing methods in diverse traffic scenarios, highlighting their generalization capabilities and reliability.
\end{itemize}
\begin{figure*}[t]
\centering
\includegraphics[width=0.95\textwidth]{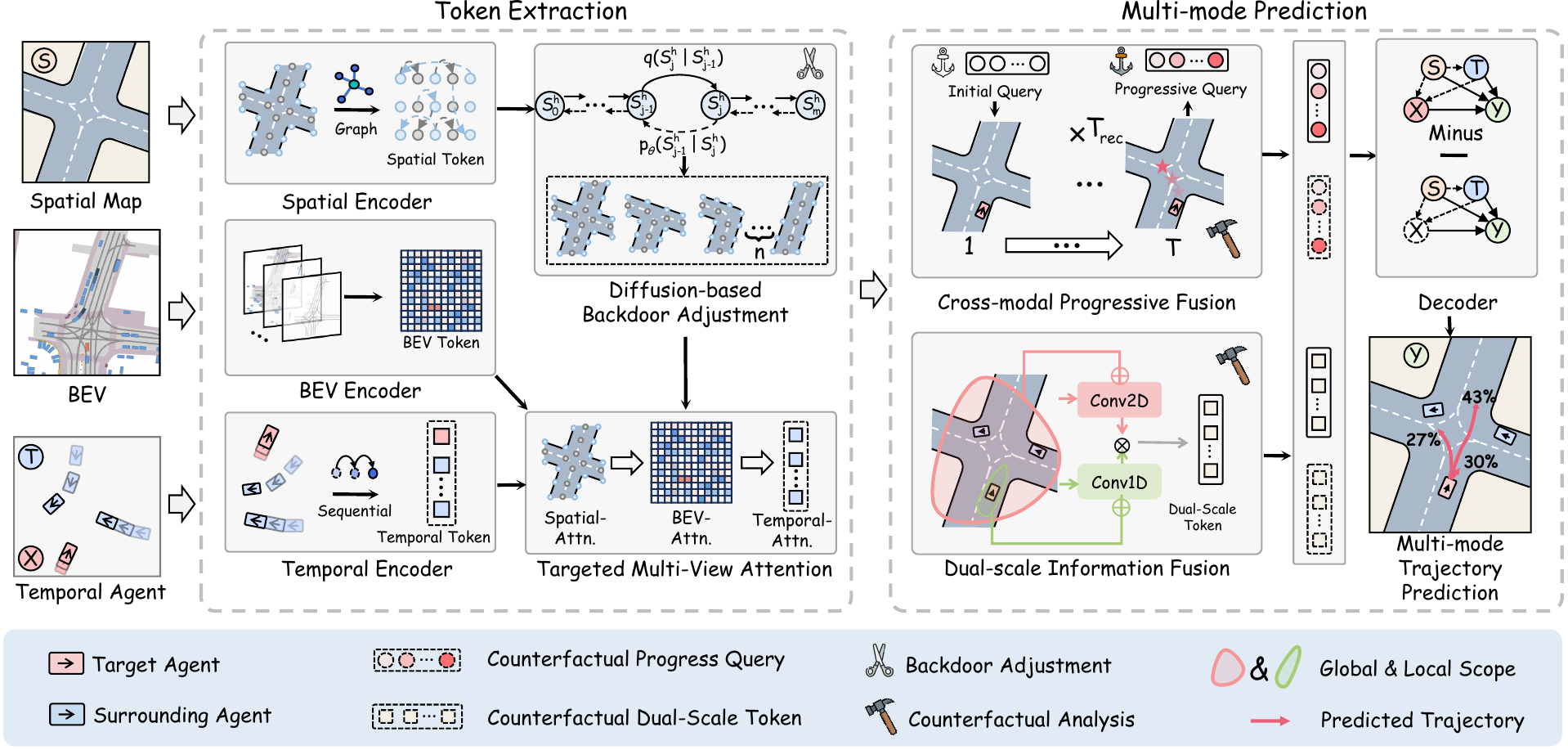}
\caption{ {Overall framework of our two-stage model. The first stage involves token extraction with spatial, BEV, and temporal encoders producing tokens \(\{S^h, B^h, T^h, X^h\}\). Spatial token \(S^h\) undergoes diffusion-based backdoor adjustment, generating \(S^{h,i}\). Combining \(S^{h,i}\) with \(\{B^h, T^h, X^h\}\) via multi-view attention yields \(X_{attn}^i\). In the second stage, \(X_{attn}^i\), initial query \(Q_0\), and counterfactual token \(X_c\) undergo cross-modal progressive fusion, producing final query \(Q^i\) and counterfactual query \(Q_c^i\). In parallel, \(X_{attn}^i\) and \(X_c\) undergo dual-scale fusion to form \(G^i\) and \(G_c^i\). The Causal Decoder then synthesizes the multi-modal predictions.}}
\label{workflow}
\end{figure*}
\section{Related Work}
\subsection{Trajectory Prediction for Autonomous Vehicles} Trajectory prediction has undergone a remarkable evolution, transitioning from traditional physics-based models and classical machine learning approaches to sophisticated deep learning architectures. Early methods, such as the Kinematic bicycle model \cite{wang2025wake} and Kalman filtering \cite{wang2023wsip} are often limited in their adaptability to complex scenarios. As a result, these methods are typically restricted to simpler environments and short-term prediction tasks. The field has seen significant progress with the rapid advancement of deep learning technologies. Recent studies have introduced sophisticated models like gated recurrent unit (GRU) \cite{li2024deep,liao2024human}, long short-term memory (LSTM) \cite{deo2018convolutional,xie2021congestion}, graph attention network (GAT) \cite{mo2022multi,liao2024less}, generative models \cite{ijcai2024p811}, and the Transformers \cite{liu2024laformer,chen2024q,liao2025safecast,liao2024mftraj}, expanding the potential for accurate long-term prediction in complex scenarios. While contemporary trajectory prediction models have demonstrated remarkable performance, their reliance on purely data-driven approaches presents fundamental limitations. These methods excel at identifying statistical regularities in training data but fail to capture the essential causal relationships that dictate agent behaviors in dynamic scenes. 
\subsection{Causal Inference in Autonomous Driving} Causal inference is a statistical framework designed to estimate the impact of one variable on another by identifying causal effects and mitigating the influence of confounding factors. In this context, two fundamental methodologies have emerged as particularly powerful for causal identification: backdoor adjustment, which enables unbiased effect estimation through proper conditioning on observed covariates, and counterfactual analysis, which examines hypothetical scenarios under different intervention conditions. The former \cite{pearl2009causality} involves identifying and controlling for confounding variables to reduce bias, while the latter \cite{kuang2020causal} evaluates causal effects by constructing hypothetical scenarios to explore ``what if'' questions. In the field of trajectory prediction, previous studies employing causal inference have respectively utilized counterfactual analysis \cite{chen2021human} and backdoor adjustment \cite{ge2023causal} to alleviate the impact of environmental confounders in traffic scenarios. However, due to an incomplete decomposition of these environmental factors, these methods often fall short in uncovering the underlying causal relationships. To address this limitation, our work advances beyond this limitation through a novel decomposition that separately models spatial and temporal environmental factors, enabling more precise confounding control. By unifying backdoor adjustment with counterfactual reasoning, we establish a more complete causal representation that preserves genuine relationships while eliminating spurious correlations, ultimately achieving superior robustness compared to existing methods.
\section{Methodology}
\subsection{Problem Formulation}
At each time step \( t \), our model utilizes historical observations from \( t-t_h \) to \( t \), denoted as \( X_0^{t-t_h:t} \), along with comprehensive traffic information--spatial map data \( S \), temporal agent data \( T \) and a BEV of the traffic scene to forecast the vehicle's trajectory over the next \( t_f \) time steps, represented as \( Y = \{X_0^{t+1:t+t_f}\} \). Specifically, \( S \) encompasses lane identifiers and precise coordinates of lane markings, while \( T \) includes the historical and predicted states of \( n \) surrounding agents, represented as \( T = \{X_{1:n}^{t-t_h:t+t_f}, p_{1:n}^{t:t+t_f}\} \). 
\subsection{A Causal View on Trajectory Prediction}
\label{causal}
\textbf{Causal Graph.}
This study constructs a causal graph to effectively map the relationships among the historical observations of the target agent \( X \), its future trajectory \( Y \), the temporal agent data \( T \), and the spatial map data \( S \). As shown in Figure \ref{fig:head} (c),  \( X \) and \( T \) influence \( Y \), represented by the edges \( X \rightarrow Y \) and \( T \rightarrow Y \). Additionally, \( S \) impacts both the historical and future trajectories, creating backdoor paths \( Y \leftarrow S \rightarrow X \) and \( Y \leftarrow S \rightarrow T \). In causal inference, variables like \( S \) and \( T \) that influence multiple other variables are termed confounders. The presence of confounders, particularly \( S \), can bias the model's ability to learn the distributional properties of \( X \) and \( T \), skewing it towards more common scenarios. For example, if straight-line trajectories dominate training data, the model may neglect specific traffic features, leading to incorrect predictions for turns. The confounding effect of \( T \) further hampers the model's ability to assess the impacts of \( T \) and \( X \) on \( Y \), causing overfitting in interactions.\\
\textbf{Casual Inference.} Given that spatial map data \( S \) is relatively fixed and observable, backdoor adjustment mitigates bias by categorizing confounding factors into distinct groups and making predictions for each group. Therefore, it is necessary to systematically enumerate \( S \) and eliminate interference in the pathways \( X \leftarrow S \rightarrow Y \) and \( T \leftarrow S \rightarrow Y \). This enables the model to capture genuine causal relationships.
\begin{equation}
\tilde{{Y}} = \sum_{i=1}^{n} g_\theta\left(X, S=s_i, T\right) P\left(s_i\right)
\label{equ:backdoor}
\end{equation}
where \( g_\theta \) is the trajectory model, \( s_i \) denotes the enumerated spatial map data, and \( P(s_i) \) is set to \( 1/n \) based on the principle of maximum entropy, with \( n \) being the assumed number of environmental categories \cite{ge2023causal}. 
Although backdoor adjustment eliminates the confounding effects of spatial map data \( S \), it does not address the complexities introduced by temporal agent data \( T \). Causal theory suggests that counterfactual analysis can mitigate these effects by intervening in the historical trajectory data \( X \) to isolate the impact of \( T \). This involves using the \( do(\cdot) \) operator to replace the factual trajectory \( X \) with a hypothetical one, fixing \( X \) to a specific value and isolating it from its antecedents. Formally,
\begin{equation}
\tilde{Y}_c = \sum_{i=1}^{n} g_{\theta} \left( do \left( X = X_c \right), S = s_i, T \right) P(s_i) \label{eq:backdoor} 
\end{equation}
where \( X_c \) represents counterfactual values. We combine backdoor adjustment and counterfactual analysis together in a comprehensive manner to derive: \( Y = \tilde{{Y}} -\tilde{{Y_c}} \).
\subsection{Proposed Model}\label{Methodology}
Figure \ref{workflow} showcases the overall pipeline of our model—token extraction, and multi-mode prediction—into a cohesive system.
This subsection outlines the structure of our model.
\subsubsection{Token Extraction}
From a causal inference perspective, predicting future trajectories requires isolating the factors influencing the target object's decisions and movements. Due to the heterogeneity of traffic information, we employ specialized encoders--Spatial Encoder, Temporal Encoder, and BEV Encoder--to extract comprehensive features from diverse data sources.\\
\textbf{Spatial Encoder.} In this encoder, the spatial map data \( S \in \mathbb{R}^{N_m \times n\times W_m} \) is tokenized into a sequence using the GRU layers. Then, GAT  is applied to extract the spatial tokens \( S^h \in \mathbb{R}^{N_m \times D}\) for the following backdoor adjustment process. Here, \( N_m \) represents the number of map polylines, \( n \) signifies the number of points within each polyline. \( w_m \) indicates the number of attributes for each point, such as location and road type, $D$ denotes the dimension after encoding.\\
\textbf{Temporal Encoder.} The GRU layers are used in this encoder to produce target  \(X^h \in \mathbb{R}^{D}\) and surrounding \(T^h \in \mathbb{R}^{N_a \times D}\) tokens for the raw observations \( X \in \mathbb{R}^{T \times W_a}\) and temporal agent data \( T \in \mathbb{R}^{N_a \times T \times W_a} \). Here, $T$ represents the number of history frames, $W_a$ is the number of state information, $N_a$ signifies the number of surrounding agents.\\
\textbf{BEV Encoder.} Recent studies have shown that explicitly modelling the heterogeneity of driving scenes by incorporating diverse modalities can substantially improve a model's ability to interpret complex interactions, thereby enhancing its accuracy. Unlike the polyline representation, the BEV format uses rasterized images to represent spatial data. Our proposed encoder hierarchically extracts frame-wise pyramid features and fuses them into BEV tokens to facilitate the interaction between these two modalities. BEV is reorganized into a three-dimensional tensor with a hierarchical structure, comprising three distinct semantic layers: Agent, Map and Raster. To mitigate the risk of occlusion that arises when directly overlaying semantic layers onto images, we employ a series of convolutional kernels with varying sizes to extract key information from these semantic layers. This allows for a comprehensive capture of the interactions among the heterogeneous elements of the scene. Finally, We apply an MLP to refine the feature, resulting in the generation of \( B^h \).\\ 
\textbf{Diffusion-based Backdoor Adjustment.} 
This component aims to mitigate the impact of confounding factors in spatial maps by generating a diverse set of potential traffic scenarios, a process known as backdoor adjustment. Since backdoor adjustment requires the stratification of \(S\), we propose leveraging spatial maps generated by a diffusion model to automatically approximate the stratified structure of \(S\). Through this approach, we effectively sever the directed edges \(S \to X\) and \(S \to T\), thereby eliminating spurious correlations and ensuring more accurate causal relationships \cite{ge2023causal}. Specifically, we utilize the output \(S^h\) from the spatial encoder as the input \(S_0^h\) for backdoor adjustment and define a diffusion sequence \((S_0^h, S_1^h, \ldots, S_k^h)\), with the forward process gradually introduces Gaussian noise to systematically disrupt the deterministic structure of the road network, eventually rendering it entirely stochastic. Formally,
\begin{equation}
q(S_j^h) = f_{\textit{noised}}\left(S_{j-1}^h\right) \text { for } j=1, \ldots, m 
\end{equation}
Conversely, the backward process employs a transformer-based architecture to iteratively reconstruct the typical road structure from this stochastic state, ultimately generating multiple instances. Mathematically,
\begin{equation}
p_\theta\left(S_{j-1}^h\right)=f_{\textit{denoised }}\left(S_j^h\right) \textit{ for } j=1, \ldots, m 
\end{equation}
By repeating this generative process \(n\) times, we obtain a backdoor spatial token set \(\bar{S} = \{ S^{h,1}, S^{h,2}, \ldots, S^{h,n} \}\). \\
\textbf{Targeted Multi-View Attention Module.} To enhance the understanding of environmental influences on the target future trajectory and facilitate causal integration, we employ a multi-view attention mechanism across spatial, BEV, and temporal contexts. Formally,
\begin{equation}
{X}_{s_{i}} = \textit{Spatial-Attn.} \left(q={X}^h, k={S}^{h,i}, v={S}^{h,i}\right)
\end{equation}
\begin{equation}
{X}_{b} = \textit{BEV-Attn.} \left(q={X}^h, k={B}^h, v={B}^h\right)
\end{equation}
\begin{equation}
{X}_{t} = \textit{Temporal-Attn.} \left(q={X}^h, k={T}^h, v={T}^h\right)
\end{equation}
The extracted tokens are processed through an aggregated MLP, producing the contextual target token \( X_\textit{attn}^i \).
\subsubsection{Multi-mode Predictions}
\textbf{Cross-modal Progressive Fusion.} This component is designed to integrate cross-modal tokens using progressive fusion. We initialize with an anchor-free query \(Q_0\) to forecast the target agent's future trajectories. In the following progressive stages, this initial query operates as an adaptive anchor point. Specifically, an attention mechanism is employed to progressively refine our adaptive anchor, identifying the agent's intended position at the current moment, which then becomes the initial query for the subsequent update. After \( T_\textit{rec} \) iterations, the definitive progressive inference anchor \( {Q}^i \) is developed. Notably, counterfactual analysis entails envisioning an alternate scenario where actual data is replaced with hypothetical data. To reduce the impact of confounding variables \( T \) in the trajectory prediction system, we substitute the historical trajectories with zero vectors and rerun the progressive fusion module to generate counterfactual anchor \( {Q}_{c}^i \).\\
\textbf{Dual-scale Information Fusion.} This component employs a CNN-based framework to extract and integrate the target agent and its surrounding agent information, resulting in the feature representation \( G^{i} \). For counterfactual analysis, the original target values are substituted with counterfactual values, and the module is applied synchronously to obtain the corresponding counterfactual representation \( G_{c}^{i} \).\\
\textbf{Causal Decoder.} In this decoder, the factual tokens \( {Q}^i \) and \( G^{i} \), along with the corresponding counterfactual tokens \( {Q}_{c}^i \) and \(  G_{c}^{i} \) are integrated using MLPs to produce composite token \( \tilde{Y} \) and \( \tilde{Y}_c \). This composite token is then processed by a GRU-based decoder, which generates the predicted probability of maneuver \( y^k_\textit{man} \) with the corresponding maneuver-based trajectory probability mixture model \( Y^k \). 
\begin{table*}[tbp]
\centering
\setlength{\tabcolsep}{3mm}
\resizebox{0.85\linewidth}{!}
{
\begin{tabular}{c|c|ccc|c|ccc}
\toprule
Model & WSADE & ADEv & ADEp & ADEb & WSFDE & FDEv & FDEp & FDEb \\
\midrule
TPNet \cite{fang2021tpnet}  & 1.2800 & 2.2100 & 0.7400 & 1.8500 & 2.3400 & 3.8600 & 1.4100 & 3.4000 \\
TP-EGT \cite{tp-egt} & 1.1900 & 2.0500 & 0.7000 & 1.7200 & 2.1400 & 3.5300 & 1.2800 & 3.1600 \\
S2TNet \cite{chen2022s2tnet}   & 1.1679 & 1.9874 & 0.6834 & 1.7000 & 2.1798 & 3.5783 & 1.3048 & 3.2151 \\
MSTG \cite{mstg}   & 1.1546 & {1.9850} & {0.6710} & {1.6745} & {2.1281} & 3.5842 & {1.2652} & {3.0792} \\
SafeCast \cite{liao2025safecast}   & {1.1253} & {1.9372} & {0.6561} & \underline{1.6247} & {2.1024} & {3.5061} & {1.2524} & {3.0657} \\
CoT-Drive \cite{liao2025cot}   & \underline{1.0958} & \underline{1.8933} & \underline{0.6179} & {1.6305} & \underline{2.0260} & \underline{3.3541} & \underline{1.1893} & \underline{3.0244} \\
 \midrule
\textbf{Ours} & \textbf{1.0681} & \textbf{1.8275} & \textbf{0.6138} & \textbf{1.5423} & \textbf{1.9174} & \textbf{3.1992} & \textbf{1.2164} & \textbf{2.6004} \\
\bottomrule
\end{tabular}}
\caption{ {Evaluation results on the ApolloScape dataset. $\text{ADE}_{v/p/b}$ and $\text{FDE}_{v/p/b}$ are the ADE and FDE for the vehicles, pedestrians, and bicycles, respectively. \textbf{Bold} and \underline{underlined} values represent the best and second-best performance in each category.}}
\label{tab:apollo}
\end{table*}
\begin{table}[tbp]
\centering
\resizebox{\linewidth}{!}
{
\begin{tabular}{ccccccc}
\toprule
\text{Model}  & {$\text{minADE}_{10}\downarrow$ } & {$\text{minADE}_5\downarrow$}  & {$\text{minADE}_1\downarrow$} & {$\text{FDE}\downarrow$} \\
\midrule
Trajectron++ \cite{salzmann2020trajectron++}  & 1.51 & 1.88 & - & 9.52 \\
MultiPath \cite{chai2019multipath}   & 1.14 & 1.44 & 3.16 & 7.69 \\
LaPred \cite{kim2021lapred}   & 1.12 & 1.53 & 3.51 & 8.12 \\
PGP \cite{deo2022multimodal}  &  \underline{0.94} & 1.27 & - & 7.17 \\
EMSIN \cite{ren2024emsin}   & - & 1.77 & 3.56 & - \\
AFormer-FLN\cite{xu2024adapting}   & 1.32 & 1.83 & - & - \\
Traj-LLM \cite{lan2024traj} & 0.99 & 1.24 & - & - \\
DEMO \cite{wang2025dynamics} & 1.04 & 1.20 & \underline{2.99} & 6.90 \\
LAformer \cite{liu2024laformer}  & \textbf{0.93} & {1.19} & - & 6.95 \\ 
NEST \cite{wang2024nest}  & - & \underline{1.18} & - & \underline{6.87} \\ 
 \midrule
\textbf{Ours}   & \textbf{0.93} & \textbf{1.17} & \textbf{2.91} & \textbf{6.71} \\
\bottomrule
\end{tabular}}
\caption{Performance comparison of various models on nuScenes dataset. \textbf{Bold} values represent the best performance in each category. ``-'' denotes the missing value.}
\label{table:performance}
\end{table}
\subsection{Learning Strategy}
The training process of our model is divided into two steps. The first step involves training the backdoor adjustment module. In the second step, with the parameters of the backdoor adjustment module fixed, we train the entire model.\\
\textbf{Step-1: Diffusion-based Backdoor Adjustment Training.}
In the initial step, we train the backdoor adjustment module using the following loss:
\begin{equation}
L_{back}(\theta) = \mathbb{E}_{\epsilon, S_0^h, j}\left\|\epsilon - \epsilon_\theta(S_j^h, j)\right\|
\end{equation}where \( \epsilon \) is sampled from a normal distribution \( \mathcal{N}(0, \mathbf{I}) \)
, and the variable \( S_j^h \) is defined as: \(
S_j^h = \sqrt{\bar{\alpha_j}} S_0^h + \sqrt{1-\bar{\alpha_j}} \epsilon\).
\textbf{Step-2: Full Model Training.}  
In the second step, we train the entire model by combining two loss functions: the intention loss (\( L_\textit{int}\)) and the trajectory loss (\( L_\textit{traj}\)). Intention loss captures the driver's operational intent, while trajectory loss ensures the accuracy of the trajectory prediction. Intention loss is calculated as follows:
\begin{equation}
L_\textit{int} = -\sum_{k=1}^K \hat{y}_{\textit{man}}^k \log \left(y_{\textit{man}}^k\right)
\end{equation}
Here, \( \hat{y}_{\textit{man}}^k \) and \( y_{\textit{man}}^k \) are the true probability and the predicted probability of maneuver \( k \), respectively. Moreover, the trajectory loss \( L_\textit{traj} \) adapts to different datasets and metrics:
\begin{equation}
L_\textit{traj} = \lambda_\textit{0} L_\textit{0} + \lambda_\textit{1} L_\textit{NLL}
\end{equation}where \( L_\textit{0} \) varies by dataset: min\(\text{ADE}_k\) for nuScenes, \(\text{WSADE}\) for ApolloScape, and \(\text{RMSE}\) for NGSIM, MoCAD, and HighD. In addition, \( L_\text{NLL} \) represents the negative log-likelihood loss, while the coefficients \( \lambda_0 \) and \( \lambda_1 \) are learnable weights for \( L_{0} \) and \( L_\text{NLL} \), respectively. Hence, the overall loss $L$  for the model is the sum of intention loss \( L_\text{int} \) and trajectory loss \( L_\text{traj} \), i.e. \(L = \lambda_\text{int} L_\text{int} + \lambda_\text{traj} L_\text{traj}\), where \( \lambda_\text{int} \) and \( \lambda_\text{traj} \) are the weights for intention loss and trajectory loss.
\begin{table}[!t]
  \centering
   \setlength{\tabcolsep}{0.5mm}
   \resizebox{1\linewidth}{!}{
    \begin{tabular}{c|cccccc}
    \bottomrule
    \multicolumn{1}{c}{\multirow{2}[4]{*}{Dataset}} & \multirow{2}[4]{*}{Model} & \multicolumn{5}{c}{Prediction Horizon (s)} \\
\cmidrule{3-7}    \multicolumn{1}{c}{} &       & 1     & 2     & 3     & 4     & 5  \\
     \hline
    \multirow{7}[3]{*}{NGSIM} 
        & WSiP \cite{wang2023wsip}& 0.56  & 1.23  & 2.05  & 3.08  & 4.34  \\
        & MHA-LSTM \cite{messaoud2020attention}& 0.41  & 1.01  & 1.74  & 2.67  & 3.83  \\
        & STDAN \cite{chen2022intention} & 0.39 & 0.96 & 1.61 & 2.56 & 3.67 \\ 
        & DACR-AMTP \cite{cong2023dacr}& 0.57  & 1.07  & 1.68  & 2.53  & 3.40 \\ 
        & GaVa \cite{liao2024human} & \underline{0.40} & \underline{0.94} & \underline{1.52} & 2.24 & 3.13 \\ 
        & HLTP++ \cite{liao2024less} & 0.46  & 0.98  & \underline{1.52}  & \underline{2.17}  & \underline{3.02}  \\
        & \textbf{Ours} & \textbf{0.32} &  \textbf{0.83}  & \textbf{ 1.47 } & \textbf{ 2.09 } & \textbf{ 2.87 } \\
    \midrule
    \multirow{7}[2]{*}{HighD} 
            & WSiP \cite{wang2023wsip}& 0.20  & 0.60  & 1.21  & 2.07  & 3.14  \\
            & MHA-LSTM \cite{messaoud2020attention}& 0.19  & 0.55  & 1.10  & 1.84  & 2.78  \\
            & GaVa \cite{liao2024human} & 0.17 & 0.24 & 0.42 & 0.86 & 1.31 \\ 
            & DACR-AMTP \cite{cong2023dacr} & 0.10  & 0.17  & 0.31  & 0.54  & 1.01  \\
            & HLTP++ \cite{liao2024less} & 0.11  & 0.17  & 0.30  & 0.47  & 0.75  \\
            & BAT \cite{liao2024bat} & \underline{0.08}  & \underline{0.14}  & \underline{0.20}  & \underline{0.44} &\underline{0.62}\\
            & \textbf{Ours} & \textbf{ 0.06 } & \textbf{0.12} &  \textbf{0.19}  & \textbf{ 0.39 } & \textbf{ 0.58 } \\
   \midrule
    \multirow{7}[2]{*}{MoCAD} 
            & CS-LSTM \cite{deo2018convolutional} & 1.45  & 1.98  & 2.94  & 3.56  & 4.49  \\
            & MHA-LSTM \cite{messaoud2020attention} & 1.25  & 1.48  & 2.57  & 3.22  & 4.20  \\
            & WSiP \cite{wang2023wsip} & 0.70  & 0.87  & 1.70  & 2.56  & 3.47  \\
            & HLTP++ \cite{liao2024less} & 0.60  & 0.81  & 1.56  & 2.40  & 3.19  \\
            & BAT \cite{ijcai2024p657} & {0.34} & \underline{0.70} & \underline{1.32} & \underline{2.01} & {2.57} \\
            & NEST \cite{wang2024nest} & \underline{0.32} & {0.75} & \textbf{1.27} & \underline{2.01} & \underline{2.42} \\
            & \textbf{Ours} & \textbf{ 0.30 } & \textbf{ 0.68 } & \textbf{ 1.27} & \textbf{ 1.97 } & \textbf{ 2.39 } \\
    \toprule
    \end{tabular}
    }
     \caption{Evaluation results for our model and  SOTA baselines in the NGSIM, HighD, and MoCAD datasets. RMSE (m) is the metric. }
 
  \label{tab:benchmark}
\end{table}
\begin{table*}[t]
  \centering
    \setlength{\tabcolsep}{4mm}
  \resizebox{0.8\linewidth}{!}{
    \begin{tabular}{ccccc}
      \toprule
      \multirow{4}{*}{Models} & \multicolumn{2}{c}{Add Noise} & \multicolumn{2}{c}{Drop Frame} \\
      \cmidrule(r){2-3} \cmidrule(l){4-5}
      & \textbf{minADE\(_5\)/FDE} & \textbf{minADE\(_5\)/FDE} & \textbf{minADE\(_5\)/FDE} & \textbf{minADE\(_5\)/FDE} \\
      & $\alpha = 8 $ & $\alpha = 16$ & $20\% $ & $40\%$ \\ 
      \midrule
          PGP \cite{deo2022multimodal}& 1.56/9.03 & 1.64/9.34 & 1.49/8.29 & 1.58/9.15 \\
          Q-EANet \cite{chen2024q} & \underline{1.44}/9.52 & \underline{1.50}/\underline{9.15} & 1.36/8.03 & 1.40/8.24 \\
          NEST \cite{wang2024nest} & 1.45/\underline{8.47} & 1.59/8.98 & 1.29/7.54 & \underline{1.37}/8.12\\
          DEMO \cite{wang2025dynamics} & 1.49/8.65 & 1.55/9.18 & \underline{1.31}/\underline{7.38} & 1.39/\underline{8.06}\\
          \midrule
          Ours &\textbf{1.32/7.64} & \textbf{1.40/8.03} & \textbf{1.23/6.89} & \textbf{1.26/7.08} \\
      \bottomrule
    \end{tabular}
  }
  \caption{{Robustness of different models on the nuScenes dataset.}}
  \label{table:robustness}
\end{table*}

\section{Experiments}
\label{Experiments}
\subsection{Experiment Setup}
The robustness and accuracy of our model are evaluated on five prominent real-world datasets, including nuScenes \cite{caesar2020nuscenes}, ApolloScape \cite{huang2018apolloscape}, MoCAD \cite{liao2024bat}, NGSIM \cite{deo2018convolutional} and HighD \cite{8569552}. In accordance with the ApolloScape and nuScenes Forecasting Challenge, as well as the seminal work \cite{liao2024bat,liao2025cot}, five metrics are employed to assess the efficacy of our model: \text{minADE}, \text{FDE}, \text{WSADE}, \text{WSFDE}, and \text{RMSE}. 
\begin{table}[!t]
  \centering
  \resizebox{\linewidth}{!}{
    \begin{tabular}{ccccc}
    \toprule
      \multirow{1}{*}{Model}  & \multicolumn{1}{c}{minADE\(_{5}\)} & \multicolumn{1}{c}{Params} & \multicolumn{1}{c}{Size} &  \multicolumn{1}{c}{Speed} \\
      \midrule
     LaPred \cite{kim2021lapred} & 	1.53	& 1.83M& 	16	& 25ms\\
     PGP \cite{deo2022multimodal} & 	1.30	& 0.42M	& 12& 	215ms\\
     Traj-LLM \cite{lan2024traj} & 	1.24	& 7.72M & 	12	& 98ms\\
     DEMO \cite{wang2025dynamics} & 	1.20	& 0.87M & 12	& 124ms\\
     LAformer \cite{liu2024laformer} & 1.19 & 2.59M & 12 & 115ms\\
     \midrule
     Ours & \textbf{1.17} & \textbf{0.28M} & 	12& 	\textbf{57ms}\\
      \toprule
    \end{tabular}
    }
       \caption{Performance comparison on nuScenes dataset.}
    \label{efficiency}
\end{table}
\begin{table}[!t]
  \centering
  \setlength{\tabcolsep}{3mm}
  \resizebox{\linewidth}{!}{
    \begin{tabular}{cccccccc}
      \toprule
      \multirow{2}[3]{*}{Component} & \multicolumn{5}{c}{Ablated variants} \\ 
      \cmidrule(r){2-6} 
      & A & B & C & D & E \\ 
      \midrule
      BEV Encoder & \ding{55} & \ding{52} & \ding{52} & \ding{52} & \ding{52} \\ 
      Human-like Progressive Fusion & \ding{52}& \ding{55} & \ding{52}  & \ding{52} & \ding{52}\\ 
      Dual-scale Information Fusion & \ding{52} & \ding{52} & \ding{55}& \ding{52} & \ding{52} \\ 
      Causal Inference & \ding{52} & \ding{52} & \ding{52} & \ding{55}& \ding{52} \\ 
      \bottomrule
    \end{tabular}
  }
  \caption{ {Different components of ablation study.}}
  \label{ablation}
\end{table}
\begin{table}[!t]
  \centering
  \setlength{\tabcolsep}{4mm}
  \resizebox{\linewidth}{!}{
    \begin{tabular}{lcccc}
      \toprule
      \multirow{2}{*}{Models} & \multicolumn{2}{c}{nuScenes} & \multicolumn{2}{c}{ApolloScape} \\
      \cmidrule(r){2-3} \cmidrule(l){4-5}
      & minADE$_{5}$ & minADE$_{10}$ & WSADE & WSFDE \\
      \midrule
      A & 1.27 & 1.12 & - & -\\
      B & 1.45 & 1.21 & 1.2870 & 2.1900\\
      C & 1.24 & 1.07 & 1.2210 & 2.0913\\
      D & 1.59 & 1.35 & 1.3264 & 2.2934\\
      E & \textbf{1.17} & \textbf{0.93} & \textbf{1.0681} & \textbf{1.9174} \\
      \bottomrule
    \end{tabular}
  }
  \caption{ {Ablation results on nuScenes and ApolloScape.}}
  \label{table:a}
\end{table}

\begin{figure}[htb]
\centering
\includegraphics[width=0.48\textwidth]{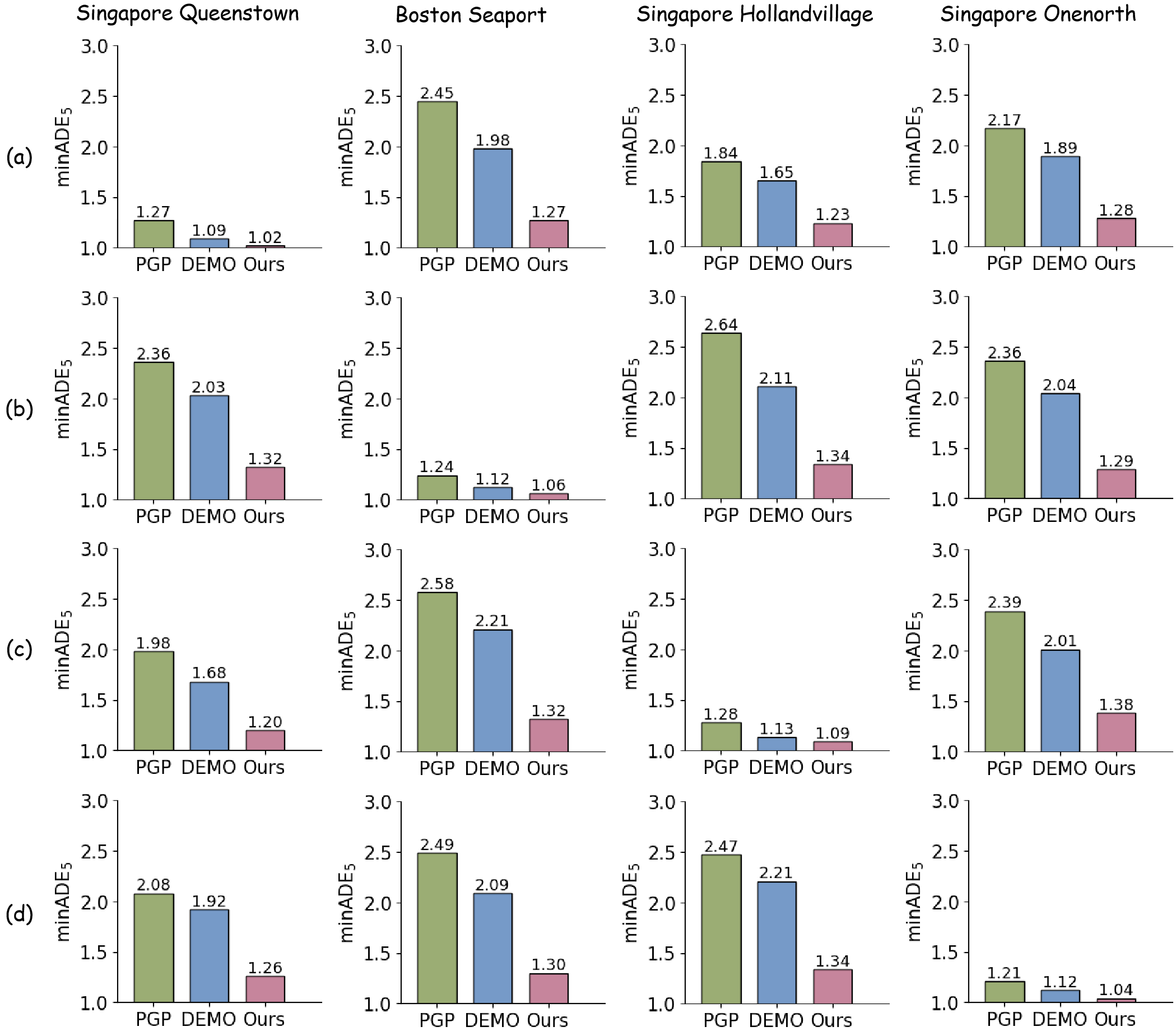}
\caption{
 {Exploration of the domain generalization ability for different models. (a) Singapore Queenstown, (b) Boston Seaport, (c) Singapore Hollandvillage, and (d) Singapore Onenorth. The results show the performance of models trained in these regions and tested in various other regions, which are evaluated using the minADE$_5$.}}
\label{fig:generalization}
\end{figure}
\begin{figure*}[t]
\centering
\includegraphics[width=0.93\textwidth]{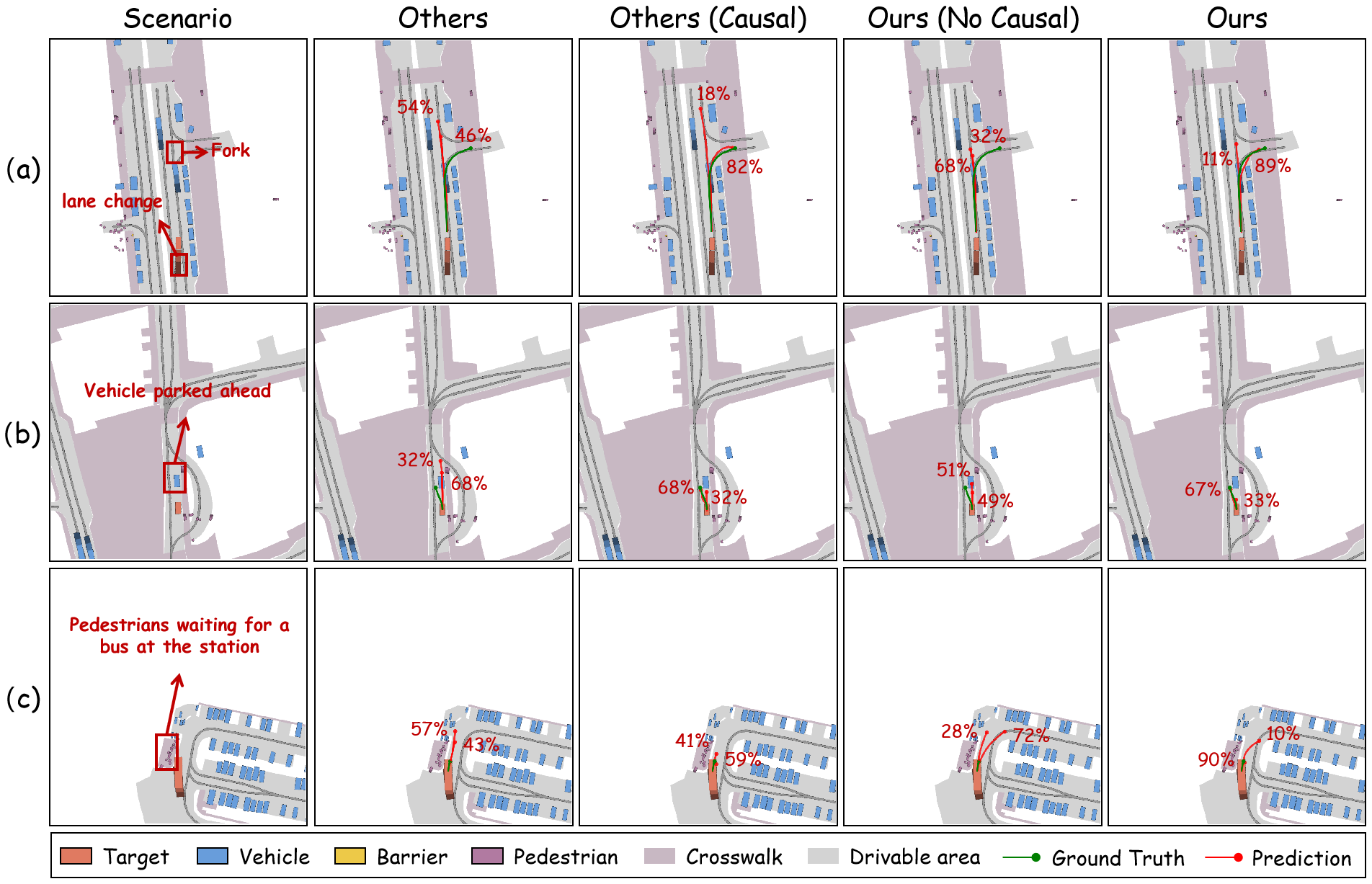}
\caption{{Comparative analysis of the impact of causal inference on our model and others across challenging scenes. For clarity, the number of predicted trajectories is set to \(k\) = 2. The probability of each predicted trajectory is shown in the subfigures.}}
\label{fig:nuscenes}
\end{figure*} 

\subsection{Evaluation Results}
\textbf{Comparison to SOTA Baselines.} 
As shown in Tables \ref{tab:apollo}, \ref{table:performance}, and \ref{tab:benchmark}, our model outperforms the current SOTA baselines across five real-world datasets. On the ApolloScape dataset, our model surpasses MSTG \cite{mstg} in WSADE and WSFDE by 1.84\% and 8.20\%, respectively. In the nuScenes benchmark, it achieves improvements of at least 7.91\% in $\text{minADE}_{1}$ and 4.00\% in $\text{FDE}$, particularly in intersections. For NGSIM and HighD datasets, it improves the long-term prediction (3s-5s) by 13.82\% and 6.45\% in RMSE, respectively. For the MoCAD dataset, which represents urban roads with right-hand drive, we see an improvement of up to 6.60\% in long-term predictions. These results validate the prediction accurancy of our model across challenging scenes, including highways, roundabouts, and congested urban roads.\\
\textbf{Comparison of Robustness.}  
We evaluate the robustness of our causal inference model using two methodologies: introducing controlled noise \cite{bagi2023generative} and randomly removing frames. To simulate observation noise, we augment the nuScenes dataset by defining a noise variable \(\sigma_t\) as a function of curvature \(\gamma_t\), formulated as follows:
\begin{equation}
   \begin{aligned}
    \gamma_t&:= \left(\dot{x}_{t + \delta_t} - \dot{x}_t \right)^2 + \left(\dot{y}_{t + \delta_t} - \dot{y}_t \right)^2, 
    \sigma_t:= \alpha \left( \gamma_t + 1 \right)
    \end{aligned} 
\end{equation}
where \(\dot{x}_t = x_{t + 1} - x_t\) and \(\dot{y}_t = y_{t + 1} - y_t\) represent agent velocities, and \(\alpha\) controls the noise magnitude. We set \(\alpha \in \{1, 2\}\) during training and \(\alpha \in \{8, 16\}\) during testing to simulate severe perturbations.
To assess robustness against missing data, 20\% or 40\% of test frames are randomly removed and replaced with zeroes. Table \ref{table:robustness} shows that under both noise and frame removal conditions, our model still outperforms SOTA baselines. These results demonstrate its ability to maintain high prediction accuracy by effectively learning and leveraging causal relationships.\\
\textbf{Comparison of Domain Generalization.}  
We partition the nuScenes dataset into four subsets and verify the statistical distinction in driving behaviors using a Kolmogorov-Smirnov test \cite{berger2014kolmogorov}, which yields p-values below the standard threshold. This confirm significant behavioral differences across regions. To assess the generalization capability of our causal inference model, we perform cross-training and testing, where the model is trained on one subset and tested on others. As shown in Figure \ref{fig:generalization}, our model achieves robust performance in unseen regions, demonstrating its effectiveness in enhancing domain generalization across diverse scenes.\\
\noindent\textbf{Comparison of Model Performance and Efficiency.} We conduct a comparative analysis of trainable parameters and inference speed on the nuScenes dataset. Table \ref{efficiency} provides detailed information on various models, including baseline data from Traj-LLM \cite{lan2024traj}, which reports the time required to predict trajectories for 12 agents using an RTX 4090 GPU. To ensure fairness, we use the same hardware and configuration to measure inference times. The results show that it not only has the smallest number of trainable parameters but also achieves an inference time of 57ms per sample, demonstrating its suitability for real-time applications.

\subsection{Ablation Study}  
We present detailed ablation variants in Table \ref{ablation} and Table \ref{table:a}. Method E represents the complete model, which achieves the SOTA performance across all metrics, demonstrating the synergistic effect of its components. Model D, which eliminates all causal analysis modules, including Backdoor Adjustment using diffusion, counterfactual analysis in progressive fusion, and dual-scale information fusion, and replaces the Differential Causal Decoder with a simple GRU, exhibits the poorest performance across most metrics, highlighting the significance of the causal inference paradigm in accurately learning casual relationship. Model B, which substitutes the progressive anchor fusion module with a single-stage anchor, also shows degraded performance, indicating the importance of progressive fusion in driving scenarios for precise trajectory prediction. The results of Method A and Method C show varying degrees of performance degradation, further confirming the necessity of each component within the model.
\subsection{Validation for Plug-and-Play Causal Inference} 
We incorporate the causal paradigm into the PGP model by replacing its GRU-based encoder with a backdoor adaptation mechanism and by integrating counterfactual reasoning into the decoder. On the nuScenes dataset, as shown in Figure \ref{fig:nuscenes}, our integrated model effectively captures causal relationships in complex lane-forking scenarios, where both the original and enhanced PGP versions fail to predict turning maneuvers. These results establish the role of causal inference in advancing behavior prediction performance.
\section{Conclusion}
\label{Conclusion}
This study advances trajectory prediction for AVs by integrating causal inference and progressive fusion strategies. Addressing the limitations of traditional data-driven methods that often neglect causal relationships, our model improves prediction accuracy and reliability through the use of causal graphs, backdoor adjustment, and counterfactual analysis, enabling it to distinguish true causal effects from correlations in complex traffic scenarios. Evaluation results on five real-world datasets show significant performance improvements across multiple metrics, marking a significant step forward in the field of trajectory prediction for fully AD systems.
\section*{Acknowledgements}
This work was supported by the Shenzhen-Hong Kong-Macau Science and Technology Program Category C [SGDX20230821095159012], University of Macau [SRG2023-00037-IOTSC, MYRG-GRG2024-00284-IOTSC], the Science and Technology Development Fund of Macau [0021/2022/ITP, 0122/2024/RIB2 and 001/2024/SKL], the State Key Lab of Intelligent Transportation System [2024-B001], and the Jiangsu Provincial Science and Technology Program [BZ2024055].
\bibliographystyle{named}
\bibliography{ijcai25}

\begin{thebibliography}{}

\bibitem[\protect\citeauthoryear{Bagi \bgroup \em et al.\egroup }{2023}]{bagi2023generative}
Shayan Shirahmad~Gale Bagi, Zahra Gharaee, Oliver Schulte, and Mark Crowley.
\newblock Generative causal representation learning for out-of-distribution motion forecasting.
\newblock In {\em ICML}, pages 31596--31612. PMLR, 2023.

\bibitem[\protect\citeauthoryear{Berger and Zhou}{2014}]{berger2014kolmogorov}
Vance~W Berger and YanYan Zhou.
\newblock Kolmogorov--smirnov test: Overview.
\newblock {\em Wiley statsref: Statistics reference online}, 2014.

\bibitem[\protect\citeauthoryear{Caesar \bgroup \em et al.\egroup }{2020}]{caesar2020nuscenes}
Holger Caesar, Varun Bankiti, Alex~H Lang, Sourabh Vora, Venice~Erin Liong, Qiang Xu, Anush Krishnan, Yu~Pan, Giancarlo Baldan, and Oscar Beijbom.
\newblock nuscenes: A multimodal dataset for autonomous driving.
\newblock In {\em CVPR}, pages 11621--11631, 2020.

\bibitem[\protect\citeauthoryear{Chai \bgroup \em et al.\egroup }{2019}]{chai2019multipath}
Yuning Chai, Benjamin Sapp, Mayank Bansal, and Dragomir Anguelov.
\newblock Multipath: Multiple probabilistic anchor trajectory hypotheses for behavior prediction, 2019.

\bibitem[\protect\citeauthoryear{Chen \bgroup \em et al.\egroup }{2021}]{chen2021human}
Guangyi Chen, Junlong Li, Jiwen Lu, and Jie Zhou.
\newblock Human trajectory prediction via counterfactual analysis.
\newblock In {\em ICCV}, pages 9824--9833, 2021.

\bibitem[\protect\citeauthoryear{Chen \bgroup \em et al.\egroup }{2022a}]{chen2022s2tnet}
Weihuang Chen, Fangfang Wang, and Hongbin Sun.
\newblock S2tnet: Spatio-temporal transformer networks for trajectory prediction in autonomous driving, 2022.

\bibitem[\protect\citeauthoryear{Chen \bgroup \em et al.\egroup }{2022b}]{chen2022intention}
Xiaobo Chen, Huanjia Zhang, Feng Zhao, Yu~Hu, Chenkai Tan, and Jian Yang.
\newblock Intention-aware vehicle trajectory prediction based on spatial-temporal dynamic attention network for internet of vehicles.
\newblock {\em IEEE Transactions on Intelligent Transportation Systems}, 23(10):19471--19483, 2022.

\bibitem[\protect\citeauthoryear{Chen \bgroup \em et al.\egroup }{2024}]{chen2024q}
Jiuyu Chen, Zhongli Wang, Jian Wang, and Baigen Cai.
\newblock Q-eanet: Implicit social modeling for trajectory prediction via experience-anchored queries.
\newblock {\em IET Intelligent Transport Systems}, 18(6):1004--1015, 2024.

\bibitem[\protect\citeauthoryear{Cong \bgroup \em et al.\egroup }{2023}]{cong2023dacr}
Peichao Cong, Yixuan Xiao, Xianquan Wan, Murong Deng, Jiaxing Li, and Xin Zhang.
\newblock Dacr-amtp: Adaptive multi-modal vehicle trajectory prediction for dynamic drivable areas based on collision risk.
\newblock {\em IEEE Transactions on Intelligent Vehicles}, 2023.

\bibitem[\protect\citeauthoryear{Deo and Trivedi}{2018}]{deo2018convolutional}
Nachiket Deo and Mohan~M Trivedi.
\newblock Convolutional social pooling for vehicle trajectory prediction.
\newblock In {\em Proceedings of the IEEE conference on computer vision and pattern recognition workshops}, pages 1468--1476, 2018.

\bibitem[\protect\citeauthoryear{Deo \bgroup \em et al.\egroup }{2022}]{deo2022multimodal}
Nachiket Deo, Eric Wolff, and Oscar Beijbom.
\newblock Multimodal trajectory prediction conditioned on lane-graph traversals.
\newblock In {\em CoRL}, pages 203--212. PMLR, 2022.

\bibitem[\protect\citeauthoryear{Fang \bgroup \em et al.\egroup }{2021}]{fang2021tpnet}
Liangji Fang, Qinhong Jiang, Jianping Shi, and Bolei Zhou.
\newblock Tpnet: Trajectory proposal network for motion prediction, 2021.

\bibitem[\protect\citeauthoryear{Ge \bgroup \em et al.\egroup }{2023}]{ge2023causal}
Chunjiang Ge, Shiji Song, and Gao Huang.
\newblock Causal intervention for human trajectory prediction with cross attention mechanism.
\newblock In {\em AAAI}, volume~37, pages 658--666, 2023.

\bibitem[\protect\citeauthoryear{Guan \bgroup \em et al.\egroup }{2024}]{guan2024world}
Yanchen Guan, Haicheng Liao, Zhenning Li, Jia Hu, Runze Yuan, Yunjian Li, Guohui Zhang, and Chengzhong Xu.
\newblock World models for autonomous driving: An initial survey.
\newblock {\em IEEE Transactions on Intelligent Vehicles}, 2024.

\bibitem[\protect\citeauthoryear{Huang \bgroup \em et al.\egroup }{2018}]{huang2018apolloscape}
Xinyu Huang, Xinjing Cheng, Qichuan Geng, Binbin Cao, Dingfu Zhou, Peng Wang, Yuanqing Lin, and Ruigang Yang.
\newblock The apolloscape dataset for autonomous driving.
\newblock In {\em CVPR workshops}, pages 954--960, 2018.

\bibitem[\protect\citeauthoryear{Kim \bgroup \em et al.\egroup }{2021}]{kim2021lapred}
ByeoungDo Kim, Seong~Hyeon Park, Seokhwan Lee, Elbek Khoshimjonov, Dongsuk Kum, Junsoo Kim, Jeong~Soo Kim, and Jun~Won Choi.
\newblock Lapred: Lane-aware prediction of multi-modal future trajectories of dynamic agents, 2021.

\bibitem[\protect\citeauthoryear{Krajewski \bgroup \em et al.\egroup }{2018}]{8569552}
Robert Krajewski, Julian Bock, Laurent Kloeker, and Lutz Eckstein.
\newblock The highd dataset: A drone dataset of naturalistic vehicle trajectories on german highways for validation of highly automated driving systems.
\newblock In {\em 2018 21st International Conference on Intelligent Transportation Systems (ITSC)}, pages 2118--2125, 2018.

\bibitem[\protect\citeauthoryear{Kuang \bgroup \em et al.\egroup }{2020}]{kuang2020causal}
Kun Kuang, Lian Li, Zhi Geng, Lei Xu, Kun Zhang, Beishui Liao, Huaxin Huang, Peng Ding, Wang Miao, and Zhichao Jiang.
\newblock Causal inference.
\newblock {\em Engineering}, 6(3):253--263, 2020.

\bibitem[\protect\citeauthoryear{Lan \bgroup \em et al.\egroup }{2024}]{lan2024traj}
Zhengxing Lan, Lingshan Liu, Bo~Fan, Yisheng Lv, Yilong Ren, and Zhiyong Cui.
\newblock Traj-llm: A new exploration for empowering trajectory prediction with pre-trained large language models.
\newblock {\em IEEE Transactions on Intelligent Vehicles}, 2024.

\bibitem[\protect\citeauthoryear{Li \bgroup \em et al.\egroup }{2024}]{li2024deep}
Can Li, Wei Liu, and Hai Yang.
\newblock Deep causal inference for understanding the impact of meteorological variations on traffic.
\newblock {\em Transportation Research Part C: Emerging Technologies}, 165:104744, 2024.

\bibitem[\protect\citeauthoryear{Liao \bgroup \em et al.\egroup }{2024a}]{ijcai2024p811}
Haicheng Liao, Xuelin Li, Yongkang Li, Hanlin Kong, Chengyue Wang, Bonan Wang, Yanchen Guan, KaHou Tam, and Zhenning Li.
\newblock Cdstraj: Characterized diffusion and spatial-temporal interaction network for trajectory prediction in autonomous driving.
\newblock In Kate Larson, editor, {\em Proceedings of the Thirty-Third International Joint Conference on Artificial Intelligence, {IJCAI-24}}, pages 7331--7339. International Joint Conferences on Artificial Intelligence Organization, 8 2024.
\newblock AI for Good.

\bibitem[\protect\citeauthoryear{Liao \bgroup \em et al.\egroup }{2024b}]{liao2024less}
Haicheng Liao, Yongkang Li, Zhenning Li, Chengyue Wang, Guofa Li, Chunlin Tian, Zilin Bian, Kaiqun Zhu, Zhiyong Cui, and Jia Hu.
\newblock Less is more: Efficient brain-inspired learning for autonomous driving trajectory prediction.
\newblock In {\em ECAI 2024}, pages 4361--4368. IOS Press, 2024.

\bibitem[\protect\citeauthoryear{Liao \bgroup \em et al.\egroup }{2024c}]{liao2024bat}
Haicheng Liao, Zhenning Li, Huanming Shen, Wenxuan Zeng, Dongping Liao, Guofa Li, and Chengzhong Xu.
\newblock Bat: Behavior-aware human-like trajectory prediction for autonomous driving.
\newblock In {\em AAAI}, volume~38, pages 10332--10340, 2024.

\bibitem[\protect\citeauthoryear{Liao \bgroup \em et al.\egroup }{2024d}]{ijcai2024p657}
Haicheng Liao, Zhenning Li, Chengyue Wang, Huanming Shen, Dongping Liao, Bonan Wang, Guofa Li, and Chengzhong Xu.
\newblock Mftraj: Map-free, behavior-driven trajectory prediction for autonomous driving.
\newblock In Kate Larson, editor, {\em IJCAI}, pages 5945--5953, 8 2024.

\bibitem[\protect\citeauthoryear{Liao \bgroup \em et al.\egroup }{2024e}]{liao2024mftraj}
Haicheng Liao, Zhenning Li, Chengyue Wang, Huanming Shen, Bonan Wang, Dongping Liao, Guofa Li, and Chengzhong Xu.
\newblock Mftraj: Map-free, behavior-driven trajectory prediction for autonomous driving.
\newblock {\em arXiv preprint arXiv:2405.01266}, 2024.

\bibitem[\protect\citeauthoryear{Liao \bgroup \em et al.\egroup }{2024f}]{liao2024human}
Haicheng Liao, Shangqian Liu, Yongkang Li, Zhenning Li, Chengyue Wang, Yunjian Li, Shengbo~Eben Li, and Chengzhong Xu.
\newblock Human observation-inspired trajectory prediction for autonomous driving in mixed-autonomy traffic environments.
\newblock In {\em 2024 ICRA}, pages 14212--14219. IEEE, 2024.

\bibitem[\protect\citeauthoryear{Liao \bgroup \em et al.\egroup }{2025a}]{liao2025safecast}
Haicheng Liao, Hanlin Kong, Bin Rao, Bonan Wang, Chengyue Wang, Guyang Yu, Yuming Huang, Ruru Tang, Chengzhong Xu, and Zhenning Li.
\newblock Safecast: Risk-responsive motion forecasting for autonomous vehicles.
\newblock {\em arXiv preprint arXiv:2503.22541}, 2025.

\bibitem[\protect\citeauthoryear{Liao \bgroup \em et al.\egroup }{2025b}]{liao2025cot}
Haicheng Liao, Hanlin Kong, Bonan Wang, Chengyue Wang, Wang Ye, Zhengbing He, Chengzhong Xu, and Zhenning Li.
\newblock Cot-drive: Efficient motion forecasting for autonomous driving with llms and chain-of-thought prompting.
\newblock {\em IEEE Transactions on Artificial Intelligence}, 2025.

\bibitem[\protect\citeauthoryear{Liao \bgroup \em et al.\egroup }{2025c}]{tsliao}
Haicheng Liao, Zhenning Li, Guohui Zhang, Keqiang Li, and Chengzhong Xu.
\newblock Toward human-like trajectory prediction for autonomous driving: A behavior-centric approach.
\newblock {\em Transportation Science}, 05 2025.

\bibitem[\protect\citeauthoryear{Liu \bgroup \em et al.\egroup }{2024}]{liu2024laformer}
Mengmeng Liu, Hao Cheng, Lin Chen, Hellward Broszio, Jiangtao Li, Runjiang Zhao, Monika Sester, and Michael~Ying Yang.
\newblock Laformer: Trajectory prediction for autonomous driving with lane-aware scene constraints.
\newblock In {\em CVPR}, pages 2039--2049, 2024.

\bibitem[\protect\citeauthoryear{Messaoud \bgroup \em et al.\egroup }{2020}]{messaoud2020attention}
Kaouther Messaoud, Itheri Yahiaoui, Anne Verroust-Blondet, and Fawzi Nashashibi.
\newblock Attention based vehicle trajectory prediction.
\newblock {\em IEEE Transactions on Intelligent Vehicles}, 6(1):175--185, 2020.

\bibitem[\protect\citeauthoryear{Mo \bgroup \em et al.\egroup }{2022}]{mo2022multi}
Xiaoyu Mo, Zhiyu Huang, Yang Xing, and Chen Lv.
\newblock Multi-agent trajectory prediction with heterogeneous edge-enhanced graph attention network.
\newblock {\em IEEE Transactions on Intelligent Transportation Systems}, 23(7):9554--9567, 2022.

\bibitem[\protect\citeauthoryear{Pearl}{2009}]{pearl2009causality}
Judea Pearl.
\newblock {\em Causality}.
\newblock Cambridge university press, 2009.

\bibitem[\protect\citeauthoryear{Ren \bgroup \em et al.\egroup }{2024}]{ren2024emsin}
Yilong Ren, Zhengxing Lan, Lingshan Liu, and Haiyang Yu.
\newblock Emsin: Enhanced multi-stream interaction network for vehicle trajectory prediction.
\newblock {\em IEEE Transactions on Fuzzy Systems}, 2024.

\bibitem[\protect\citeauthoryear{Salzmann \bgroup \em et al.\egroup }{2020}]{salzmann2020trajectron++}
Tim Salzmann, Boris Ivanovic, Punarjay Chakravarty, and Marco Pavone.
\newblock Trajectron++: Dynamically-feasible trajectory forecasting with heterogeneous data.
\newblock In {\em ECCV}, pages 683--700. Springer, 2020.

\bibitem[\protect\citeauthoryear{Tang \bgroup \em et al.\egroup }{2023}]{mstg}
Luqi Tang, Fuwu Yan, Bin Zou, Wenbo Li, Chen Lv, and Kewei Wang.
\newblock Trajectory prediction for autonomous driving based on multiscale spatial-temporal graph.
\newblock {\em IET Intelligent Transport Systems}, 17(2):386--399, 2023.

\bibitem[\protect\citeauthoryear{Wang \bgroup \em et al.\egroup }{2023}]{wang2023wsip}
Renzhi Wang, Senzhang Wang, Hao Yan, and Xiang Wang.
\newblock Wsip: wave superposition inspired pooling for dynamic interactions-aware trajectory prediction.
\newblock In {\em AAAI}, volume~37, pages 4685--4692, 2023.

\bibitem[\protect\citeauthoryear{Wang \bgroup \em et al.\egroup }{2025a}]{wang2025wake}
Chengyue Wang, Haicheng Liao, Zhenning Li, and Chengzhong Xu.
\newblock Wake: Towards robust and physically feasible trajectory prediction for autonomous vehicles with wavelet and kinematics synergy.
\newblock {\em IEEE Transactions on Pattern Analysis and Machine Intelligence}, 2025.

\bibitem[\protect\citeauthoryear{Wang \bgroup \em et al.\egroup }{2025b}]{wang2024nest}
Chengyue Wang, Haicheng Liao, Bonan Wang, Yanchen Guan, Bin Rao, Ziyuan Pu, Zhiyong Cui, Chengzhong Xu, and Zhenning Li.
\newblock Nest: A neuromodulated small-world hypergraph trajectory prediction model for autonomous driving.
\newblock {\em AAAI}, 2025.

\bibitem[\protect\citeauthoryear{Wang \bgroup \em et al.\egroup }{2025c}]{wang2025dynamics}
Chengyue Wang, Haicheng Liao, Kaiqun Zhu, Guohui Zhang, and Zhenning Li.
\newblock A dynamics-enhanced learning model for multi-horizon trajectory prediction in autonomous vehicles.
\newblock {\em Information Fusion}, page 102924, 2025.

\bibitem[\protect\citeauthoryear{Xie \bgroup \em et al.\egroup }{2021}]{xie2021congestion}
Xu~Xie, Chi Zhang, Yixin Zhu, Ying~Nian Wu, and Song-Chun Zhu.
\newblock Congestion-aware multi-agent trajectory prediction for collision avoidance.
\newblock In {\em 2021 ICRA}, pages 13693--13700. IEEE, 2021.

\bibitem[\protect\citeauthoryear{Xu and Fu}{2024}]{xu2024adapting}
Yi~Xu and Yun Fu.
\newblock Adapting to length shift: Flexilength network for trajectory prediction.
\newblock In {\em CVPR}, pages 15226--15237, 2024.

\bibitem[\protect\citeauthoryear{Yang \bgroup \em et al.\egroup }{2024}]{tp-egt}
Biao Yang, Fucheng Fan, Rongrong Ni, Hai Wang, Ammar Jafaripournimchahi, and Hongyu Hu.
\newblock A multi-task learning network with a collision-aware graph transformer for traffic-agents trajectory prediction.
\newblock {\em IEEE Transactions on Intelligent Transportation Systems}, pages 1--14, 2024.

\end{thebibliography}
\end{document}